\documentclass[pmlr,twocolumn]{jmlr}

\usepackage{longtable}

\usepackage{booktabs}
\usepackage[load-configurations=version-1]{siunitx} 

 \usepackage{multirow}

\theorembodyfont{\upshape}
\theoremheaderfont{\scshape}
\theorempostheader{:}
\theoremsep{\newline}

\jmlrvolume{193}
\firstpageno{1}

\jmlryear{2022}
\jmlrworkshop{Machine Learning for Health (ML4H) 2022}

\title[Neurodevelopmental Phenotype Prediction]{Neurodevelopmental Phenotype Prediction:\titlebreak A State-of-the-Art Deep Learning Model}

\author{\Name{Dániel Unyi} \Email{unyi.daniel@tmit.bme.hu} \\
\Name{Bálint Gyires-Tóth} \Email{toth.b@tmit.bme.hu} \\
\addr Department of Telecommunications and Media Informatics \\
Budapest University of Technology and Economics \\
Műegyetem rkp. 3., H-1111 Budapest, Hungary}

\begin{document}

\maketitle

\begin{abstract}

A major challenge in medical image analysis is the automated detection of biomarkers from neuroimaging data. Traditional approaches, often based on image registration, are limited in capturing the high variability of cortical organisation across individuals. Deep learning methods have been shown to be successful in overcoming this difficulty, and some of them have even outperformed medical professionals on certain datasets. In this paper, we apply a deep neural network to analyse the cortical surface data of neonates, derived from the publicly available Developing Human Connectome Project (dHCP). Our goal is to identify neurodevelopmental biomarkers and to predict gestational age at birth based on these biomarkers. Using scans of preterm neonates acquired around the term-equivalent age, we were able to investigate the impact of preterm birth on cortical growth and maturation during late gestation. Besides reaching state-of-the-art prediction accuracy, the proposed model has much fewer parameters than the baselines, and its error stays low on both unregistered and registered cortical surfaces.

\end{abstract}


\section{Introduction}

The goal of neuroimaging is collecting visual data of the nervous system, and analysing this data in order to better understand its structure and function. High-resolution images are primarily acquired through magnetic resonance imaging (MRI), which allows non-invasive and non-radioactive measurements on the patient. Large, publicly available MRI datasets and the corresponding processing pipelines \citep{hcp0, hcp} enable the research community to work on novel machine learning models, aimed to improve our understanding of pathological conditions, and to assist medical experts in diagnosis, prognosis and therapeutic suggestions.

Deep learning models have two crucial advantages over other machine learning models: 1) they can automatically identify the most important biomarkers, automating the meticulous work known as feature selection 2) they can exploit geometric priors, like the spatial correlation of 2D/3D MRI images or surfaces. With state-of-the-art accuracy, novel deep learning methods provide valuable contributions to the field of neuroimaging, for instance the detection of neurodegenerative diseases \citep{alzheimer, s2cnn_alzheimer}, segmentation of brain tumors \citep{tumor}, or parcellation of the cerebral cortex \citep{gdl}. Some of the methods are even able to outperform medical professionals on certain datasets \citep{expert}.

The cerebral cortex is a thin, highly convoluted neural tissue, constituting the outer layer of the brain. Billions of neurons live within its folds and grooves, facilitating complex cognitive processes which ultimately make us humans and ourselves. Due to its vital role and complexity, it is intensively studied by the machine learning research community. The most common approaches for processing cortical MRI images are convolutional neural networks \citep{resnet}, and more recently, vision transformers \citep{vit}. Other approaches involve the extension of convolutions to manifolds \citep{gdl0}, based on the reconstruction of the cortical surface as a polygon mesh.

In this paper, we extend a previous line of work, organised around the Developing Human Connectome Project (dHCP). We apply a deep neural network to identify neurodevelopmental biomarkers, and based on these biomarkers, to predict scan age and birth age of preterm and term neonates. During late gestation, rapid cortical growth and maturation occurs. Since preterm birth interrupts essential cellular and molecular processes, it increases the risk of developing neuropsychiatric conditions in later life. The main contributions of this work are the following:
\begin{itemize}
    \item Our deep learning model achieves state-of-the-art accuracy in scan age prediction and birth age prediction.
    \item Using SHAP values, we investigate which myelinated cortical areas has the most significant effect onto the model's final predictions.
    \item We describe the slight modifications we made to our model, to win an international machine learning challenge in this topic.
\end{itemize}

\section{Related Work}

The cerebral cortex is the highly convoluted outer layer of the brain, often reconstructed from structural MRI scans as a polygon mesh. The deep learning based analysis of polygon meshes requires the generalisation of convolution from Euclidean grids to manifolds, which led to the development of geometric deep learning (gDL) and intrinsic mesh CNNs \citep{gdl0}.

Many previous works applied gDL techniques directly on the mesh-based representation of the cortical surface. \citet{cucurull} framed cerebral cortex parcellation as a mesh segmentation task. By utilizing graph neural networks \citep{chebnet, gat}, they segmented the Broca area of 100 participants from the Human Connectome Project, and significantly improved the Dice score over algorithmic baselines and multilayer perceptrons (which processed the cortical features independently, just as we did in this paper). A similar approach was taken by \citet{gopinath} to parcellation: the aligned spectral embedding of the cortical meshes were concatenated to the other cortical features, before feeding them into a graph neural network. They segmented the whole cortical surface of 101 subjects from the Mindboggle dataset, and achieved better Dice score than the FreeSurfer algorithm \citep{freesurfer}, one of the most popular automatic parcellation tools.

Others prefer to inflate the cortical meshes and project them onto a sphere, which is commonly referred to as "sphericalisation" \citep{preprocess}. There are various gDL approaches explicitly designed for spherical manifolds \citep{s2cnn, deepsphere, spherical_unet}. According to \citet{s2cnn_alzheimer}, spherical CNNs show superior performance to traditional CNNs in classifying Alzheimer’s disease versus cognitively normal, on 589 patients from the ADNI-1 cohort. \citet{spherical_unet} proposed Spherical U-Net, by replacing all operations in the standard U-Net \citep{unet} with equivalent spherical operations. They demonstrated the capability of their model not only in parcellation, but in predicting the cortical feature development of 370 infants in their first year after birth.

Recently, a large body of work was organised around the Developing Human Connectome Project (dHCP) - a publicly available dataset of cortical surfaces, acquired from preterm and term neonates. \citet{gdl} benchmarked a collection of gDL methods, comparing them with each other and ResNet \citep{resnet} on sphericalised cortical surfaces. They aimed to predict the postmenstrual age (PMA) at scan and gestational age (GA) at birth. On registered surfaces, the encoder of Spherical U-Net \citep{spherical_unet} performed the best, but this performance drastically dropped on unregistered surfaces. On unregistered surfaces, the rotationally equivariant Spherical CNN \citep{s2cnn} and MoNet \citep{monet} were the best performing models. The authors briefly investigated whether sphericalisation holds any disadvantage, but the corresponding results were not conclusive. In the follow-up work by \citet{sit}, they adapted vision transformers to sphericalised cortical surfaces (SiTs), representing data as a sequence of triangular patches. SiTs scored similarly or better than gDL methods in both prediction tasks, and could gain benefit from pretraining on ImageNet or on the same dataset using self-supervised learning (e.g. masked patch prediction). To encourage the endeavours of developing novel methods regarding GA regression at birth, the authors organised a \href{https://slcn.grand-challenge.org/}{MICCAI 2022 Challenge} with deadline of 8th July, 2022, leading up to our current paper.

\begin{table*}
\caption{Dataset description.}
\begin{center}
    \begin{tabular}{|c|c|c|c|c|c|}
        \hline
        \multirow{2}{*}{Dataset} &
        \multirow{2}{*}{Total} &
        \multicolumn{2}{c|}{Maturity} &
        \multicolumn{2}{c|}{Sex} \\
        \cline{3-6}
        \multirow{2}{*}{} &
        \multirow{2}{*}{} &
        Preterm &
        Term &
        Male &
        Female \\
        \hline
        PMA at scan & 530 & 111 & 419 & 288 & 242 \\
        \hline
        GA at birth & 514 & 95 & 419 & 283 & 231 \\
        \hline
    \end{tabular}
\end{center}
\label{table:dataset}
\end{table*}

\section{Methods}

\subsection{Data}

Our initial dataset was derived from the publicly available third release of the \href{http://www.developingconnectome.org/}{Developing Human Connectome Project} (dHCP). The reconstructed T1 and T2-weighted magnetic resonance images \citep{mri1, mri2, mri3} were processed according to the dHCP structural pipeline\footnote{\url{https://github.com/BioMedIA/dhcp-structural-pipeline}} \citep{preprocess}, which performed tissue segmentation \citep{tissue1, tissue2, tissue3}, surface extraction \citep{surface}, and inflation and projection to a sphere. This process also generates cortical feature descriptors, including three descriptors of cortical geometry (sulcal depth, curvature, thickness) and the T1w/T2w ratio, highly correlated to cortical myelination \citep{hcp}.

\begin{figure*}
    \centering
    \includegraphics[width=\textwidth]{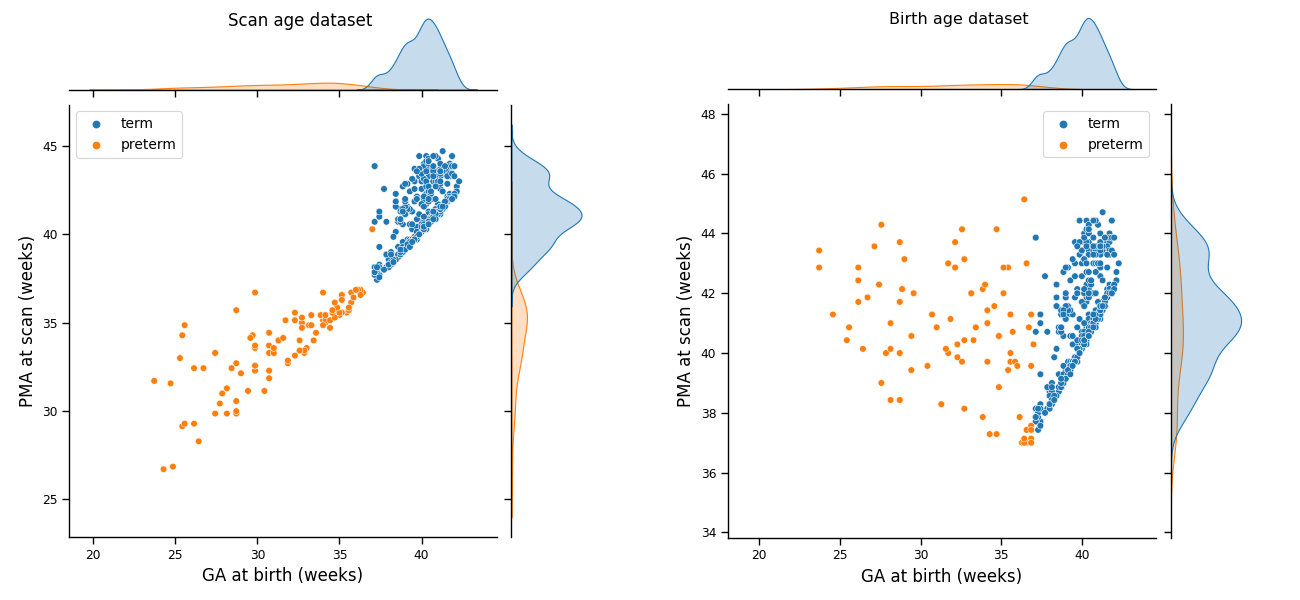}
    \caption{Distribution of the scan age and birth age datasets.}
    \label{fig:dataset}
\end{figure*}

The dataset contains the cortical surface of 588 neonatal subjects, with accompanying metadata such as gestational age (GA) at birth, postmenstrual age (PMA) at scan, sex, birthweight, head circumference, and radiology score. Table \ref{table:dataset} and Figure \ref{fig:dataset} show insight into the dataset for both tasks. Some of the preterm neonates were scanned twice: once around birth and once at term-equivalent age. Since our goal is to predict scan age as a marker of healthy development, and to explore the impact of preterm birth on term-equivalent development, we excluded the later preterm scans from scan age regression and the earlier preterm scans from birth age regression (also important for the direct comparison with previous methods).

We preprocessed the initial dataset as follows. For experiments in native space (i.e. unregistered data), we used each subject's native sphere, while in template space (i.e. registered data), we used the transform from each subject's native surface to the dHCP symmetric 40-week surface template. The 4 cortical features (sulcal depth, curvature, thickness, and T1w/T2w ratio) were resampled from the native/template sphere to the 6-th order icosphere with barycentric interpolation, implemented by the Human Connectome Project workbench software \citep{workbench}. This pipeline results 40962 equally spaced vertices per subject, with 4 cortical features per vertex. Right hemispheres were mirrored onto the sagittal plane, such that their orientation matched with left hemispheres. We computed the mean and standard deviation of the 4 features in the train set, and standardized the input data every time before training or inference.

\subsection{Model}
\begin{figure}
    \centering
    \includegraphics[width=0.4\textwidth]{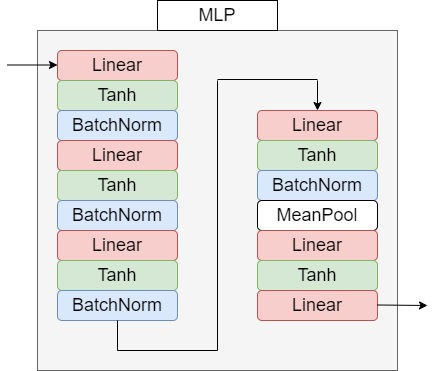}
    \caption{Our proposed architecture.}
    \label{fig:arch}
\end{figure}

We used a multilayer perceptron (MLP) in each of our experiments. The architecture is shown on Fig. \ref{fig:arch}. We applied the following layer sequence four times: \(\{Linear() \rightarrow Tanh() \rightarrow BatchNorm()\}\). After that, we performed mean pooling along the vertex dimension, and added a regression head \(Linear() \rightarrow Tanh() \rightarrow Linear()\) to produce the final predictions. Each hidden layer had 16 units, resulting fewer than 1400 trainable parameters. We had 4 input features in scan age prediction, but 5 in birth age prediction, since in this case we also incorporated scan age as a confound. The model was trained using mean squared error (MSE) as loss function, and Adam as optimizer with learning rate 0.001. Batch size was set to 32. Training was halted when the validation loss had not decreased in the last 200 epochs, and the model with the lowest validation loss was restored.

\section{Results and Discussion}

\subsection{State-of-the-Art Results}

For each of the four tasks (scan age prediction and birth age prediction in both native and template space), we trained an MLP model four times with different initial weights, and monitored the mean absolute error (MAE) in weeks\footnote{Code to reproduce the results is publicly available at \url{https://github.com/daniel-unyi-42/Neurodevelopmental-Phenotype-Prediction}}. For direct comparison with previous models \citep{gdl, sit}, we report the best MAE score achieved on the test set, alongside the standard deviation across all four trainings, in Table \ref{table:results}. We compared our MLP to the following geometric deep learning methods: S2CNN \citep{s2cnn}, ChebNet \citep{chebnet}, GConvNet \citep{gconvnet}, Spherical U-Net \citep{spherical_unet}, MoNet \citep{monet}, and the following surface vision transformers: SiT-0 was not pretrained, SiT-1 was pretrained on ImageNet, and SiT-2 was pretrained via masked patch prediction on the dHCP dataset \citep{sit}. Overall, our MLP model consistently outperformed the other models in native space, with MAE of 0.54 weeks in scan age prediction and 1.08 weeks in birth age prediction. It was also competitive compared to the best performing models in template space, with the exception of Spherical U-Net in birth age prediction, which still leads the race by a large margin. Incorporating scan age as a confound in birth age prediction yielded lower error, in agreement with the findings of \citet{sit}. We also performed 10-fold cross validation, and report the mean and standard deviation across the 10 different test folds in Table \ref{table:results}. We think this is a fairer way to evaluate and compare future models trained on this dataset.

It is important to note that our model has less than 1400 trainable parameters, a fraction of the several million parameters in previous models. We attributed the success of our lightweight MLP model to two reasons: 1)~it can generalise better on unseen data 2)~it was easier to optimise its hyperparameters (for instance, it could not train as well with asymmetric activation functions as with tanh). However, it is possible that message passing models inherently overfit this dataset: changing the first 4 linear layers to the convolutional layers of ChebNet, GConv-Net, or MoNet results an immediate drop in performance, regardless of the small number of parameters. Using an NVIDIA TITAN V GPU, our model requires only an average of 834 ms training time per epoch and 76 ms inference time.

\begin{table*}
\caption{Best \textbf{M}ean \textbf{A}bsolute \textbf{E}rror and standard deviation across 4 runs are reported (in weeks). Our MLP model achieves better results than surface vision transformers \citep{sit} and geometric deep learning methods \citep{gdl}, with only a fraction of trainable parameters. We also report the result of 10-fold cross validation in the last row.}
\begin{center}
    \begin{tabular}{|c|c|c|c|c|c|c|}
        \hline
        \multirow{2}{*}{\textbf{Method}} &
        \multicolumn{3}{c|}{\textbf{Scan age}} &
        \multicolumn{3}{c|}{\textbf{Birth age}} \\
        \cline{2-7}
        \multirow{2}{*}{} &
        \textbf{Template} &
        \textbf{Native} &
        \textbf{Avg} &
        \textbf{Template} &
        \textbf{Native} &
        \textbf{Avg} \\
        \hline \hline
        S2CNN & 0.63±0.02 & 0.73±0.25 & 0.68 & 1.35±0.68 & 1.52±0.60 & 1.44 \\
        ChebNet & 0.59±0.37 & 0.77±0.49 & 0.68 & 1.57±0.15 & 1.70±0.36 & 1.64 \\
        GConvNet & 0.75±0.13 & 0.75±0.26 & 0.75 & 1.77±0.26 & 2.30±0.74 & 2.04 \\
        Spherical UNet & 0.57±0.18 & 0.87±0.50 & 0.72 & \textbf{0.85±0.17} & 2.16±0.57 & 1.51 \\
        MoNet & 0.57±0.02 & 0.61±0.05 & 0.59 & 1.44±0.08 & 1.58±0.06 & 1.51 \\
        SiT-0 & 0.60±0.02 & 0.76±0.03 & 0.68 & 1.14±0.12 & 1.44±0.03 & 1.29 \\
        SiT-1 & 0.59±0.03 & 0.71±0.02 & 0.65 & 1.15±0.05 & 1.69±0.03 & 1.42 \\
        SiT-2 & 0.55±0.04 & 0.63±0.06 & 0.59 & 1.13±0.02 & 1.47±0.08 & 1.30 \\
        \hline\hline
        \textbf{MLP(ours)} & \textbf{0.50±0.05} & \textbf{0.54±0.01} & \textbf{0.52} & 1.08±0.05 & \textbf{1.08±0.17} & \textbf{1.08} \\
        \textbf{MLP-cv(ours)} & 0.55±0.05 & 0.55±0.03 & 0.55 & 1.00±0.12 & 1.12±0.22 & 1.06 \\
        \hline
    \end{tabular}
\end{center}
\label{table:results}
\end{table*}


\begin{figure*}[!htbp]
    \centering
    \includegraphics[width=\textwidth]{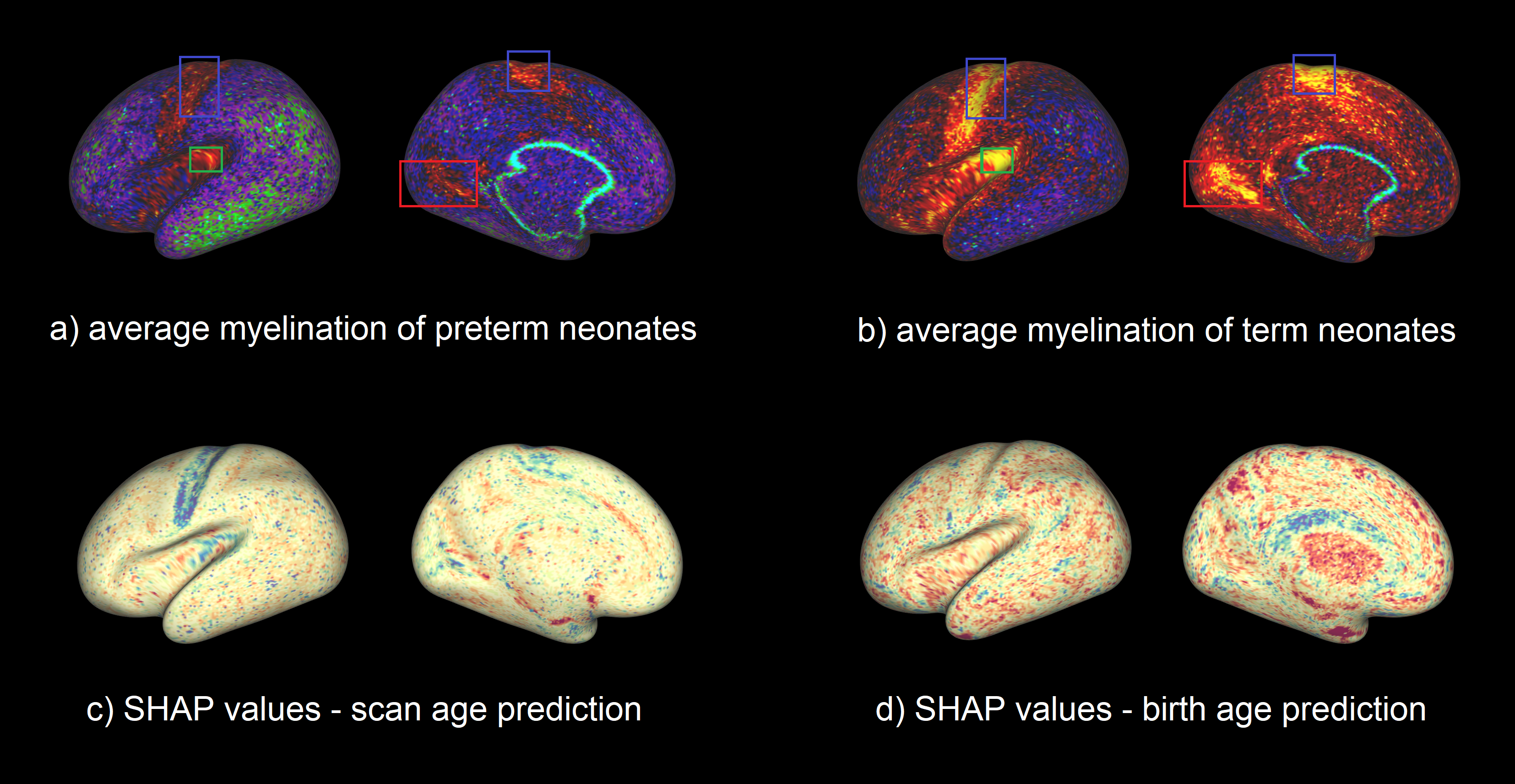}
    \caption{a) and b) show the average cortical myelination of preterm and term neonates. The increase of myelin levels can be seen in the visual (red box), auditory (green box), and somatosensory (blue box) areas. c) and d) highlight the SHAP values in scan age and birth prediction, with red and blue colours indicating that the given area is important in the model's prediction making.}
    \label{fig:shap}
\end{figure*}

\subsection{Identifying Neurodevelopmental Biomarkers}

Since our model was successful at detecting biomarkers, it is worth to investigate how it concluded the predicted values of scan age and birth age. For this purpose, we used the DeepExplainer method \citep{shap} of the SHapley Additive ExPlanations (SHAP) library\footnote{\url{https://github.com/slundberg/shap}} \citep{shap0}. SHAP assigns a so-called SHAP value to each feature, indicating its importance in the model's prediction making.

We averaged the T1w/T2w ratio of preterm (GA\textless 37 weeks) and term (GA\textgreater 37 weeks) neonates separately over the validation and test sets, to visualise the dependence of cortical myelination on maturity. Figure \ref{fig:shap} highlights the difference of myelin levels in the visual, auditory and somatosensory areas between the two gestational periods. SHAP values indicate that the latter two areas were indeed important indicators in scan age prediction, marking the process of healthy development. In birth age prediction, other parts of the cortex become more important, including the superior frontal gyrus, the praecuneus, or the temporal pole. These areas are responsible for high-level cognitive functions, like certain aspects of consciousness, memory, or visuospatial processing, and are likely impacted by preterm birth.

\subsection{SLCN: A MICCAI 2022 Challenge}

Our solution ranked first on the MICCAI 2022 Challenge called \href{https://slcn.grand-challenge.org/}{Surface Learning for Clinical Neuroimaging}. The goal of this challenge was to elicit submissions of explainable machine learning methods, which are able to identify neurodevelopmental biomarkers from cortical surface data. Thereby, submissions were evaluated on the task of birth age regression. The same train and validation sets (acquired from the dHCP) were provided to the participants that we used throughout this paper, but the test set was kept hidden during the challenge. Metadata was also provided for subjects in the train and validation sets: the PMA at scan, sex, birthweight, and head circumference. Since the PMA at scan was not available for test set subjects, we had to work out a less straightforward deconfounding strategy. The key observation was that predicting the confound and incorporating it into the loss function consistently decreased the error on the test set. Predicting the birthweight variable and incorporating it into the loss also yielded better results: the standard deviation decreased among validation errors (i.e. evaluations of the same model with different initial weights). This is understandable, since it is strongly correlated with our target variable. In conclusion, the only modification we made to our MLP model was adding 3 output units instead of 1. These 3 output units correspond to the GA at birth, the PMA at scan, and the birthweight of the subject; and since we incorporated all 3 variables into the loss function, we required the MLP model to predict all of them as accurately as possible. Our best test MAE was 1.226 in template space and 1.386 in native space.

\section{Conclusion}

In this work, we applied a deep neural network to predict neurodevelopmental phenotypes, based on the cortical surface data of preterm and term neonates. We achieved state-of-the-art results compared to previous baselines, demonstrating that models with fewer priors and fewer parameters can be more effective on certain datasets. We also identified some myelinated cortical areas that are biomarkers of healthy development, and others that are possible biomarkers of preterm birth.

Regarding future research, we plan to dissipate the debate around whether or not sphericalisation leads to the loss of useful information (as we mentioned, results in \citet{gdl} are not conclusive). We are going to apply intrinsic mesh CNNs on the high-resolution, native cortical meshes, and identify the cortical areas whose geometry may affect the predictions. We also plan to explore further metadata like biological sex, utilizing it both as confound and as target variable.

\acks{The work reported in this paper, carried out at BME, has been supported by the European Union project RRF-2.3.1-21-2022-00004 within the framework of the Artificial Intelligence National Laboratory.}

\bibliography{unyi22}






\end{document}